\documentclass[sn-mathphys,Numbered]{sn-jnl}

\usepackage[inkscapelatex=false]{svg}
\usepackage{subfig}
\usepackage{adjustbox}
\usepackage{graphicx}%
\usepackage{multirow}%
\usepackage{amsmath,amssymb,amsfonts}%
\usepackage{amsthm}%
\usepackage{mathrsfs}%
\usepackage[title]{appendix}%
\usepackage{xcolor}%
\usepackage{textcomp}%
\usepackage{manyfoot}%
\usepackage{booktabs}%
\usepackage{algorithm}%
\usepackage{algorithmicx}%
\usepackage{algpseudocode}%
\usepackage{listings}%
\graphicspath{{figures/}}

\usepackage{etoolbox}
\makeatletter
\patchcmd{\maketitleAuthor}{\fnmfont{\theAuthor@fnm} \snmfont{\theAuthor@snm}}{\fnmfont{\theAuthor@fnm}~\snmfont{\theAuthor@snm}}{}{}
\makeatother

\begin{document}

\title[Article Title]{Graph Neural Network for Stress Predictions in Stiffened Panels Under Uniform Loading}



\author[1]{Yuecheng Cai}
\author[1,2]{Jasmin Jelovica}


\affil[1]{\orgdiv{Department of Mechanical Engineering}, \orgname{The University of British Columbia}, \orgaddress{ \city{Vancouver},  \state{BC}, \country{Canada}}}

\affil[2]{\orgdiv{Department of Civil Engineering}, \orgname{The University of British Columbia}, \orgaddress{ \city{Vancouver},  \state{BC}, \country{Canada}}}



\abstract{
Machine learning (ML) and deep learning (DL) techniques have gained significant attention as reduced order models (ROMs) to computationally expensive structural analysis methods, such as finite element analysis (FEA). Graph neural network (GNN) is a particular type of neural network which processes data that can be represented as graphs. This allows for efficient representation of complex geometries that can change during conceptual design of a structure or a product. In this study, we propose a novel graph embedding technique for efficient representation of 3D stiffened panels by considering separate plate domains as vertices. This approach is considered using Graph Sampling and Aggregation (GraphSAGE) to predict stress distributions in stiffened panels with varying geometries. A comparison between a finite-element-vertex graph representation is conducted to demonstrate the effectiveness of the proposed approach. A comprehensive parametric study is performed to examine the effect of structural geometry on the prediction performance. Our results demonstrate the immense potential of graph neural networks with the proposed graph embedding method as robust reduced-order models for 3D structures.}




\maketitle

\section{Introduction}\label{sec1}
\subsection{Motivation}\label{subsec1}

The progress in developing efficient modern structures is significantly driven by advancements in structural analysis methods such as Finite Element Analysis (FEA) \cite{clough1990original}. With the adoption of this powerful tool, engineers can improve design of engineering structures. People can now routinely analyze structures with diverse geometries without the need for intricate mathematical or analytical solutions to governing differential equations. Consequently, there is a growing emphasis on pursuing cost-effective structural solutions, as they not only enhance performance but also lead to reductions in weight and manufacturing costs \cite{sullivan2018effect}.
 
Many thin-walled structures, such as bridges, ships, and aircraft incorporate stiffened panels in their design, which are generally effective in carrying a diverse range of loads and relatively easy to manufacture. The stiffened panel consists of a plate and discrete stiffeners welded onto it. The geometry of both stiffeners and face plates used in the construction is often rectangular because this is relatively easy to fabricate and can always provide good performance. The panels often need to be optimized in design. 

The optimization of stiffened panels and large structures incorporating the is mostly performed by using sizing optimization \cite{samanipour2020adaptive,jelovica2022improved} and sometimes topology optimization \cite{chu2021design}. Structural optimization often relies on FEA to estimate stresses under loads. Although FEA can be successfully applied to solve numerous complex problems, several challenges remain in structural optimization:

\begin{itemize}
  \item The computational costs increase as the complexity of the model rises, including the refinement of the mesh;
  \item Evaluation of each design requires independent simulations;
  \item Re-meshing is often necessary when the structure changes geometrically.
\end{itemize}

With these challenges, the use of FEM in structural optimization is expensive. In order to overcome this problem, traditional reduced-order models (ROMs) have been used, such as Multivariate Adaptive Regression Splines (MARS), Kriging (KRG), Radial Basis Functions (RBF), and Response Surface Method (RSM) \cite{chen2006review}. These methods aim to maintain the precision of high-fidelity models while having a relatively low computational cost.

However, applying the ROMs above to complicated engineering problems is limited due to their inherent assumptions. Moreover, they may lose fidelity when the structure changes geometrically. More recently, with the advancements in machine learning (ML) and in particular deep learning (DL) methodologies, there is a tendency to adopt ML/DL models as ROMs, taking advantage of their versatility in capturing various data properties. Particularly in mechanical engineering, ML/DL techniques are effective modelling tools and approximators that often exceed conventional ROM techniques in accuracy and capacity to represent even nonlinearities of engineering problems \cite{mai2022robust,shojaeefard2013modelling,kabir2021failure}. 

\subsection{Deep learning-based reduced order models for structural analysis}\label{subsecLiterature}

Typical ROMs require a structure to be represented parametrically, where the structural variables are identified as inputs to the ML model. The typical technique for reduced-order modelling is artificial neural networks (ANNs). One of the most commonly used is multi-layer perceptron, also known as MLP, which has been widely implemented in various fields, as evidenced by numerous scholarly works. This is because MLP with one hidden layer is a universal approximator, which can accurately predict any continuous function arbitrarily well if there are enough hidden units, as stated in Ref \cite{hornik1989multilayer}. 

An early study of MLP-based ROM for structural optimization can be found in Ref \cite{papadrakakis1998structural}, where MLP with one hidden layer was employed as ROM to replace the structural analysis in optimization. With the demand for higher prediction accuracy, deeper neural networks have been employed by researchers to accomplish more complex tasks such as the prediction of axial load carrying capacity \cite{bisagni2002post}, buckling load \cite{sun2021prediction}, shear \cite{limbachiya2021application} and lateral-torsional \cite{ferreira2022lateral} resistance, etc. More recent advancements in ANNs for structural analysis can be found in Ref. \cite{mandal2021application,zarringol2023artificial,zhu2023artificial}.

However, limited by the monolithic structure of traditional neural networks, even if additional layers can be added to increase the architecture’s complexity, MLPs are inefficient at accurately describing complicated structural behaviour since they demand a large amount of training data and computational resources. In addition, MLPs tend to overfit, and they are less interpretable than other types of neural network (NN) approaches, therefore its capabilities are limited and not suited for advanced problems. 

Therefore, to capture more complex features, some researchers have used more advanced NNs such as convolutional neural networks (CNNs) as ROMs for the structures that can be represented as images (2D matrices) or composition of images (3D matrices). A few examples can be found from Refs. \cite{ramkumar2021unconventional,banga20183d}, where different composite materials' modulus, strength, and toughness are predicted by feeding enough composite configurations (around 1,000,000). In addition, researchers utilized CNN for stress predictions in different structures, for instance, the maximum stress of brittle materials \cite{wang2021stressnet}, and stress contour in components of civil engineering structures \cite{bolandi2022deep}.

Instead of approximating the mechanical properties and structural responses of structures, some researchers also train CNNs or generative adversarial networks (GANs) to predict the optimal structural design directly; interested readers could check Refs. \cite{nie2021topologygan,achour2020development,mao2020designing,shu20203d} for more information.

However, modelling engineering structures using fixed-size vectors or matrices, such as images, proves to be challenging, considering that one of the variables is structural geometry. It introduces a dimensional change in the design input, which is primarily handled by re-training for  NN approaches such as MLP and CNN. Furthermore, stiffened panels are discrete structures, which consist of repeated structural units, e.g., stiffeners and plates, whose connection can not be neglected as they affect the mechanical response of the structure. Motivated by these two reasons, this paper proposes an approach that transforms structural models into graphs. This transformation permits flexibility in varying the dimension of design inputs, a characteristic that aligns well with the capabilities of graph neural networks (GNNs). 

GNNs have made significant strides across various fields, such as computer vision \cite{xu2017scene}, traffic prediction \cite{yao2018deep}, chemistry \cite{gilmer2017neural}, biology \cite{fout2017protein}, and recommender systems \cite{ying2018graph}, etc. In fluid mechanics, researchers have employed GNNs in replacing the expensive CFD simulations using dynamic graphs in recent years \cite{lino2022multi,shao2023pignn,pfaff2020learning}, where they converted mesh to the vertex and edge features, consequently applying GNNs to learn the temporal evolution of the properties in a fluid system. However, since GNN just gained significant attention in recent years, not many advancements have been made in the structural engineering field. In a recent study by Zheng et al. \cite{zheng2023tso}, a graph embedding approach was employed to represent 2D and 3D trusses as graphs, with vertices denoting the joints and edges representing the bars. This method aimed to optimize the arrangement of bars within trusses while adhering to stress and displacement constraints. Similarly, other recent investigations applying GNNs to truss-related problems can be found in references such as \cite{bacciu2020gentle} and \cite{whalen2022toward}. These studies employed GNNs in conjunction with various techniques, including Q-learning and transfer learning, to address distinct objectives. Nevertheless, there has been a notable absence of research dedicated to the application of GNNs to more intricate 3D structures beyond the scope of truss problems.

In this paper, we extend the previous research by introducing an innovative graph embedding technique tailored for 3D stiffened panels. We use this novel graph embedding method for predicting stresses in stiffened panels, employing a Graph Sampling and Aggregation network (GraphSAGE). We compare the proposed graph embedding with the conventional finite-element-vertex embedding that has been used in fluid mechanics. Additionally, we conduct a comprehensive parametric study for various structural geometry parameters including various boundary conditions and complex geometric variations.

\section{Methodolgy}\label{sec2}
\subsection{Basis of Graph Neural Networks}\label{subsec2}
Graph Neural Networks (GNNs) represent a specialized subset of neural networks (NNs) renowned for their capacity to handle data with graph embeddings. Unlike the CNN, which is typically used for tasks involving data that is defined on a regular grid, such as images, GNNs are designed to process and analyze data represented in the form of graphs, such as data with complex relationships between entities or data that has a natural representation as a network.

In general, a graph can be defined as $G = (V,E,A)$, where $V$ represents the set of vertices, $E$ indicates the set of edges between these vertices, and $A$ is the adjacency matrix. We denote the edge going from vertex $v_i$ to vertex $v_j$ as $e_{ij} = (v_i,v_j)\in E$. If a graph is undirected, every two vertices contain two edges $e_{ij} = (v_i,v_j)\in E$, and $e_{ji} = (v_j,v_i)\in E$. The adjacency matrix $A\in \mathbb{R}^{N\times N}$ is a convenient way to represent the graph structure, where $N=|V|$ is the number of vertices, $A_{ij} = 1$ if $e_{ij}\in E$. For an undirected graph, $A_{ij} = A_{ji}$. Therefore, a graph is associated with graph attributes $X\in \mathbb{R}^{N\times D}$ and $A\in \mathbb{R}^{N\times N}$, where $D$ is the number of input features of each vertex. It is worth mentioning that in this study, all graphs are defaulted as undirected graphs. 

To conduct convolution on a graph, researchers developed techniques such as the graph spectral method, which leverages the eigenvalues and eigenvectors of the graph Laplacian matrix to define graph convolutions, few of the most famous networks are Spectral CNN \cite{bruna2013spectral}, Chebyshev Spectral CNN (ChebNet) \cite{defferrard2016convolutional}, and Graph Convolution Network (GCN) \cite{kipf2016semi}, etc. However, they require a large memory to conduct the graph convolution, which is not efficient in handling large graphs. Hence, the emergence of the graph spatial method, which operates by leveraging the spatial interconnections among vertices and their adjacent neighbours, for instance, original Graph Neural Network (GNN) \cite{scarselli2008graph}, Graph Sampling and Aggregation (GraphSAGE) \cite{hamilton2017inductive}, and Graph Isomorphism Network (GIN) \cite{xu2018powerful}, etc.

\begin{figure}[ht]%
\centering
\includegraphics[width=0.9\textwidth]{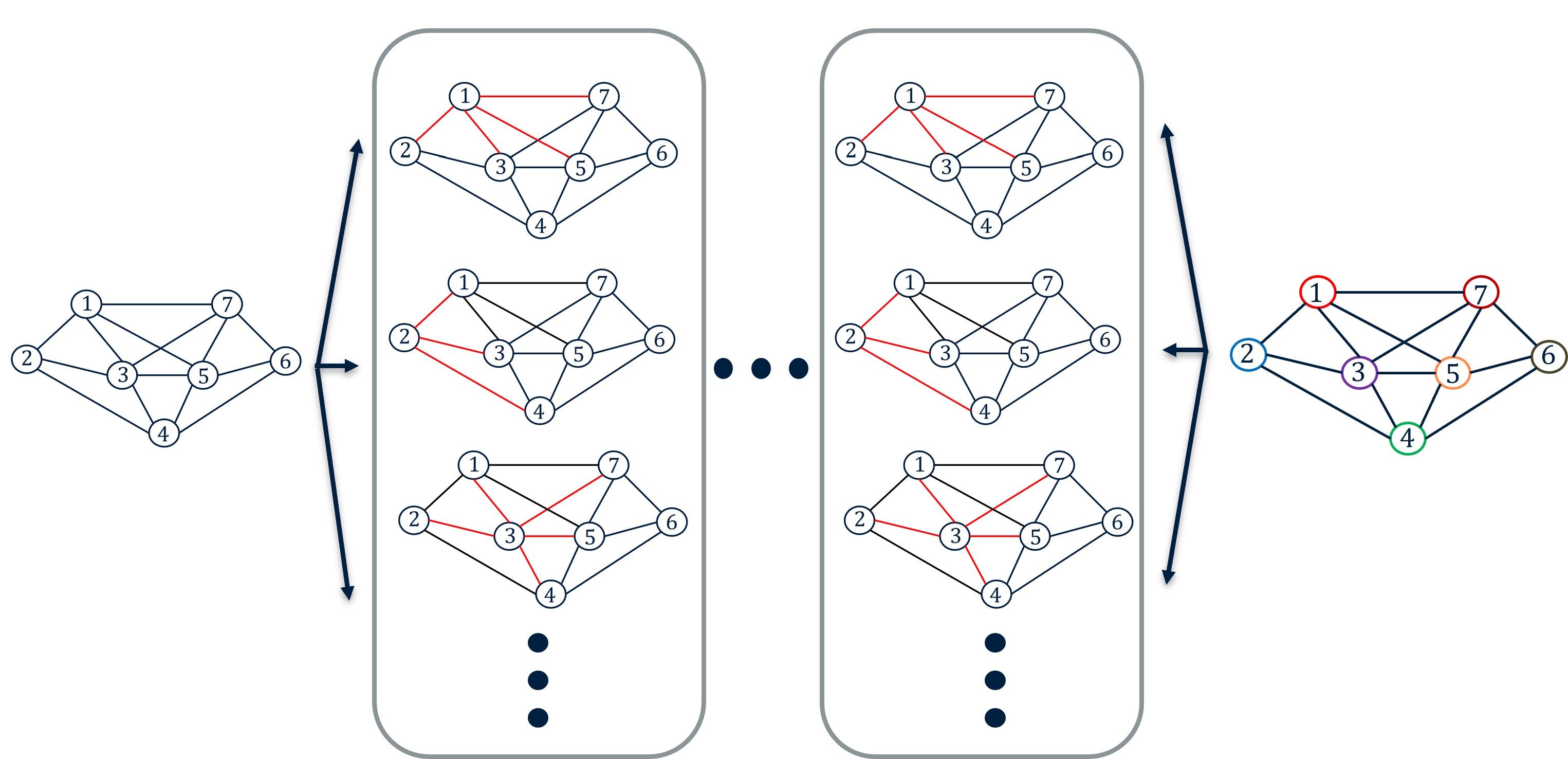}
\caption{Architecture of the GraphSAGE Network employed in this study. Each vertex represents a structural unit, including plate span, stiffener web and flange. In the input graph, vertices hold the geometric details of their respective structural units. Meanwhile, in the output graph, vertices contain the stress information associated with their corresponding structural components.}\label{fig: GraphSAGE flowchart}
\end{figure}

All the above-mentioned techniques can be summarized into message-passing neural networks (MPNNs) \cite{battaglia2018relational}, which represent a general framework of graph convolution. Under this framework, we specifically employ the GraphSAGE network \cite{hamilton2017inductive}, which is a prominent and benchmark method for many graph-related tasks. To preserve the maximum information of each vertex, the `sum' operator is determined as the aggregation function \cite{xu2018powerful}. Utilizing the MPNN framework, the GraphSAGE operator with a `sum' aggregator is defined as: 
\begin{equation}
\textbf{h}_{v}^{l}=\sigma (\textbf{W}^{l} (\textbf{h}_{v}^{l-1}+sum_{u\in \mathcal{N}(v)} \textbf{h}_{u}^{l-1}))\label{Eq: GraphSAGE}
\end{equation}
where $\textbf{h}_{v}^{l-1}$ and $\textbf{h}_{v}^{l}$ are the embedding for vertex $v$ at layer $l-1$ and $l$, respectively. $\textbf{W}^{l}$ is the trainable parameter at the current layer $l$. Message propagation is conducted through vector concatenation, followed by the message update phase through the $tanh$ activation function $\sigma$. 

A general structure of a GraphSAGE network can be found in Fig. \ref{fig: GraphSAGE flowchart}. Structural geometric data and external loading are initially transformed into the proposed graph representation, as detailed in Section \ref{subsecEmbedding}. At each layer, the GraphSAGE operator is employed to process and learn the features of the graph. Batch normalization has been utilized after each GraphSAGE convolution layer to stabilize the training process of the GNN. Mean square error (MSE) is adopted as the loss function in this study. The hyperparameters for the utilized model have been fine-tuned and are detailed in Table \ref{tabl: NN hyperparemeter}. Note that the hyperparameter settings are consistent throughout the investigation in Section \ref{sec4}.

\begin{table}[ht]
\caption{Hyperparameter setting of the Employed GraphSAGE Network}\label{tabl: NN hyperparemeter}%
\begin{tabular}{@{}llll@{}}
\toprule
Category & Value  \\
\midrule
Number of layers    & 32   \\
Number of hidden neurons for each layer   & 64  \\
Activation function    & tanh   \\
Optimizer   & adam \\
Learning rate  & 0.02  \\
Batch size  & 512 \\
L2 regularization factor  & 1e-4\\
\botrule
\end{tabular}
\end{table}

\subsection{Proposed graph embedding for stiffened panels}\label{subsecEmbedding}

As indicated in the previous section, graph embedding is the prerequisite for a structure being handled by a GNN. In this study, we propose a simple but efficient method to graphically represent the stiffened panel. 

\begin{figure}[ht]%
\centering
\includegraphics[width=0.7\textwidth]{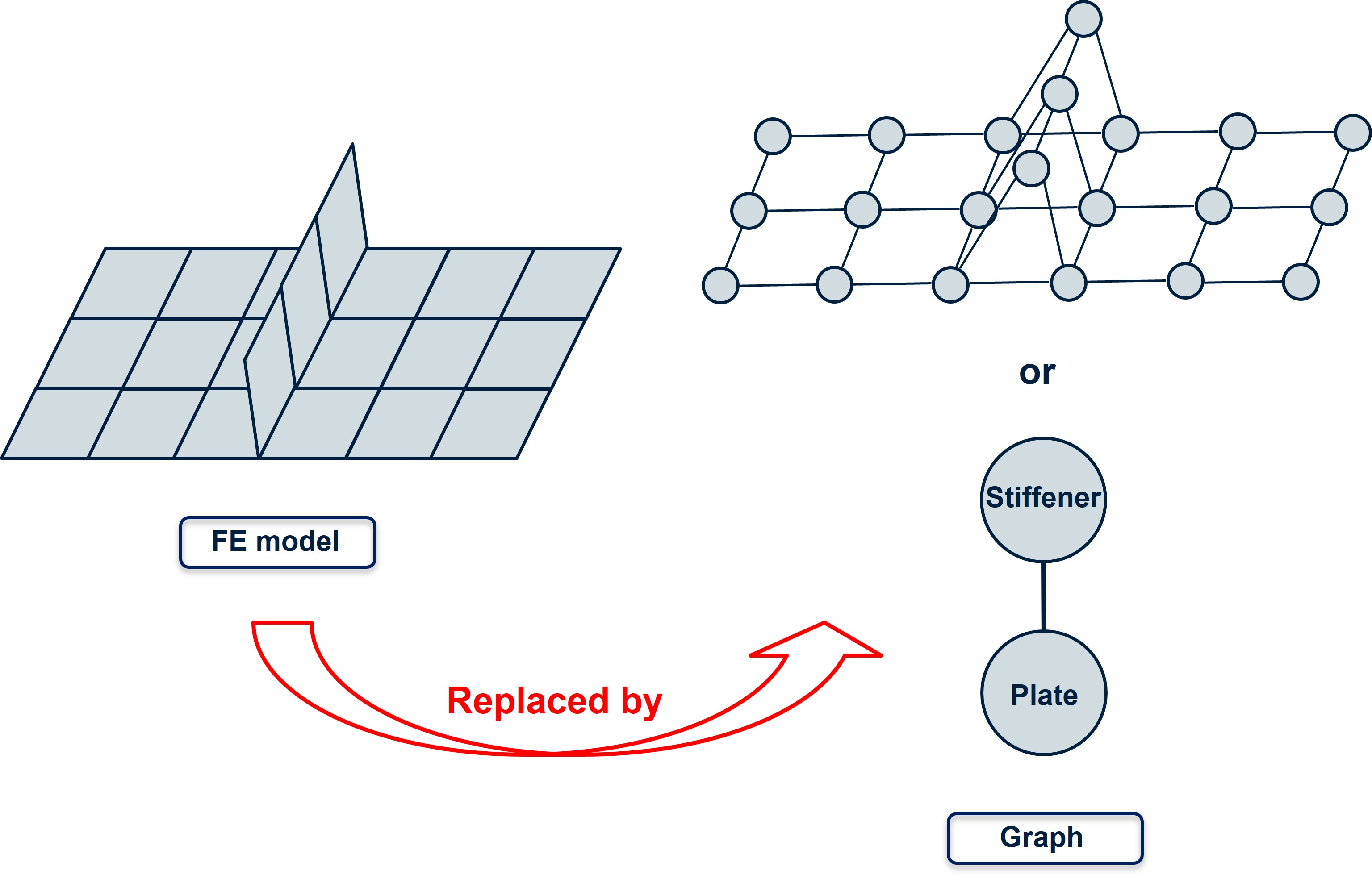}
\caption{Proposed graph representation versus finite element-vertex graph representation. The representation of the finite element-vertex graph is displayed on the top right. Each element is substituted by a vertex, and the connections between elements are denoted as edges. The proposed graph layout is presented at the bottom. Every vertex stands for a structural component, such as the plate span, stiffener web, and flange, with their interrelationships defined by edges.
}\label{SP graph representation}
\end{figure}

Fig. \ref{SP graph representation} illustrates a conventional finite-element-vertex graph representation where each element is a vertex. Our approach is to represent each stiffener or plate between stiffeners as a vertex. The conventional method maps each finite element to a corresponding vertex, while connections between these elements are determined as edges. This transformation allows the direct representation of the finite element (FE) model of the structure as a graph. Consequently, as the mesh refinement progresses, the graph naturally scales in size in a proportional manner.

However, it is essential to consider that the time complexity of GNN training is proportional to the number of vertices $|V|$. The employed graphs in this study are sparse graphs, with the following assumptions:
\begin{itemize}
  \item The GNN architecture is fixed, with a depth of $L$, and a transformed graph embedding size of $F_i$ at layer $i$;
  \item The average number of neighbours $k$ of vertices in the graph is about the same for different graph representations;
  \item The time complexity of feature transformation and aggregation dominates other operators.
\end{itemize}
The time complexity for feature transformation is $\mathcal{O}(N \times \Sigma_{i=1}^L F_{i-1} \times F_i)$, and $\mathcal{O}(N \times k \times \Sigma_{i=1}^L F_{i})$ for aggregation. Combining the time complexity for both feature transformation and aggregation, we can get the total training complexity:
\begin{equation}
\mathcal{O}(N \times \Sigma_{i=1}^L F_{i-1} \times F_i) + \mathcal{O}(N \times k \times \Sigma_{i=1}^L F_{i})\label{Eq: TimeComplexity}
\end{equation}
Given constants $C_{1} = \Sigma_{i=1}^L F_{i-1} \times F_i$ and $C_{2} = k \times \Sigma_{i=1}^L F_{i}$, the combined complexity can be written as: $\mathcal{O}(N \times (C_{1} + C_{2}))$. We can clearly see that $\mathcal{O}(N \times (C_{1} + C_{2})) \propto \mathcal{O}(N)$, where the primary factor is the number of vertices $N$ in a graph. Therefore, adopting the conventional finite-element-vertex graph representation can lead to a significant increase in computational resources as the level of structural discretization grows.

In this paper, we introduce an efficient graph embedding technique designed for modelling stiffened panels with a reduced number of vertices, as indicated in Figure \ref{SP graph representation}. More specifically, we represent each structural unit, including plates between stiffeners, stiffener webs, and flanges, as individual vertices within the graph. Each vertex incorporates both geometrical information of the physical entity and its boundary conditions. Our objective is to reduce the total number of vertices from $N$ to a smaller number $N^{\prime}$, thereby expediting the GNN's training process by approximately $N/N^{\prime}$ than with the conventional approach. For each vertex, we employ eight variables, including structural unit width, length, thickness, and boundary conditions for each edge, together with the value of the applied pressure. The connectivity between each structural unit is encoded by the adjacency matrix and is not reflected in the vertex input embedding. 

The objective of this study is to predict the von Mises stress distribution across stiffened panels, which is crucial for structural design. Thus, it is necessary to obtain the grid space (mesh) stress information of each structural unit. Utilizing the proposed graph embedding, and given that the vertex output dimensions do not affect the GNN training time, we have allocated a grid space of $10\times 20$ to each structural unit. For each vertex, the von Mises stress information of each node on the grid space is determined as output. Since vertices in the graph can only accept vector inputs, the number of input features $D$ for each vertex has been reshaped to a dimension of $1 \times 200$.

\section{Data preparation}\label{sec3}

The case study are stiffened panels because they represent the basic repeating unit of many large-scale structures. To approximate the stiffened panel of real-life structures, the span of the panel has been set as $3 m\times 3 m$. Main plate thickness ranges from $10mm$ to $20mm$. Each panel contains $2$ to $8$ stiffeners, with a random height from $0.1m$ to $0.4m$. All stiffeners have a T-shaped cross-section, with a rectangular flange whose width ranges from $0.05m$ to $0.15m$. The thickness for both stiffener web and flange can change from $5mm$ to $20mm$. We allow the stiffener/flange height and thickness to continuously vary in this range, which allows a wider design space. We assume that the plate of the panel is subjected to a uniformly distributed pressure, which ranges from $0.05MPa$ to $0.1MPa$. The summary of the upper and lower limits for the structure's geometrical details can be found in Table \ref{tabl: var limit}. 

\begin{table}[ht]
\caption{Lower and upper limits of stiffened panel geometrical variables}\label{tabl: var limit}%
\begin{tabular}{@{}llll@{}}
\toprule
Category & Lower limit  & Upper limit & Unit\\
\midrule
Plate thickness    & 10   & 20  & mm  \\
Stiffener thickness    & 5   & 20  & mm  \\
Stiffener height    & 100   & 400  & mm  \\
Flange thickness  & 5  & 20 & mm \\
Flange width   & 50 & 150 & mm \\
Number of stiffeners  & 2 & 8 & $-$ \\
\botrule
\end{tabular}
\end{table}

In this study, we have systematically considered two key variables for a comprehensive parametric study: structural boundary conditions and geometric variations. In both cases, we maintain the remaining variables fixed to carefully assess the effect of introducing a specific variable under investigation. For each case separately we prepare a total of 2000 randomly generated design configurations, allocating 80\% of them for training and 20\% for validation. Detailed analysis for each parametric test case is exhibited and discussed in Section \ref{sec4}.

The dataset is obtained using parametric models prepared in MATLAB and executed through ABAQUS FEM software. The static analysis is performed on a model discretized using the `SR4' element. The training procedure of GNN is executed using Pytorch Geometric and carried out on a computing device with a GTX 3090 GPU.

\section{Results and discussion}\label{sec4}

In this section, we first compare the proposed entity-vertex graph embedding with the finite element-vertex embedding, to demonstrate the efficiency of the proposed approach. Afterwards, we present the results of our investigation into the influence of various parameters on the accuracy of the trained GNN model. The parameters under examination include boundary conditions and geometric complexity. Understanding how these parameters affect the neural network's performance is crucial in optimizing its accuracy and robustness for real-world applications. For each parameter, we first discuss its influence on neural network accuracy, followed by a detailed analysis and comparison of predictions with the ground truth data.

\subsection{Impact of graph representation on GNN computational resources}

To quantitatively study the influence of graph embedding on GNN training time complexity as discussed in \ref{subsecEmbedding}, we compare our proposed stiffened panel graph embedding with a finite-element-vertex graph representation through a simple test case. This comparison primarily focuses on the training time difference between the two graph embedding techniques, the test case presented here omits the stiffener flange for simplicity. All other structural variables remain consistent with those described in Table \ref{tabl: var limit}. Two graph neural networks have been trained independently based on the same dataset, GraphSAGE architecture, and hyperparameter setting, differing only in graph embedding. An example for both embeddings can be found in Fig. \ref{SP graph representation}. 

\begin{figure}[H]
    \centering
    \subfloat[\normalsize Element-vertex embedding]{\includegraphics[width=0.3\textwidth]{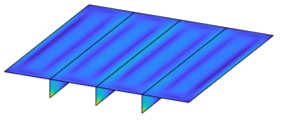}\label{fig: comparison ele}}
    \hfill
    \subfloat[\normalsize Entity-vertex embedding]{\includegraphics[width=0.3\textwidth]{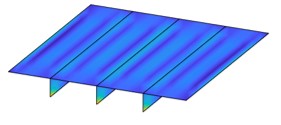}\label{fig: comparison entity}}
    \hfill
    \subfloat[\normalsize Ground truth]{\includegraphics[width=0.3\textwidth]{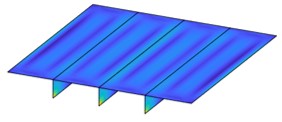}\label{fig: comparison true}}
    \caption{Comparison between finite element-vertex embedding and the proposed entity-vertex embedding}\label{fig comparison}
\end{figure}

Figure \ref{fig comparison} illustrates the comparison of stress contour predictions using GraphSAGE with two graph embeddings, both of which yield comparable performance levels. The training computational resources for both approaches are presented in Table \ref{tabl:embedding comparison}. Utilizing the same batch size, the GNN trained with the proposed graph embedding achieves approximately 27 times faster training times per epoch compared to the conventional method. The GPU memory requirements also differ significantly: the proposed approach consumes 0.5 GB for a batch size of 64, while the finite element-vertex embedding requires 23.4 GB. These results demonstrate the benefits of the proposed approach, which is useful across a broader spectrum of computer devices.

\begin{table}[ht]
\caption{Graph embedding computational resources comparison}\label{tabl:embedding comparison}%
\begin{tabular}{@{}lcc@{}}
\toprule
Category & Entity-vertex embedding  & Element-vertex embedding \\
\midrule
Running time per epoch    & 0.2553 seconds   & 6.9375 seconds    \\
GPU memory    & 0.5 GB   & 23.4 GB    \\
\botrule
\end{tabular}
\end{table}

\subsection{Parametric study}\label{subsecParametricStudy}

\subsubsection{The effect of boundary conditions}\label{subsecBC}

In this subsection, we examine the impact of structural boundary conditions on the performance of the GraphSAGE model. Entity-vertex embedding is used, where the entity is plate between the stiffeners (or edge and stiffener), stiffener web and stiffener flange. While considering all the variables outlined in Table \ref{tabl: var limit}, we also incorporate the structural boundary conditions as an additional variable. Specifically, plate edges and stiffeners are assigned with random boundary conditions separately. All edges of the plate are either simply supported or fixed. In addition, the edges of the stiffener web and flange are free, simply supported, or fixed.

\begin{table}[ht]
\centering
\caption{Comparison of GraphSAGE predictions and FEA results for stiffened panel with varying structural boundary conditions}\label{tabl: Case 1 contour}%
\begin{tabular}{@{}ccccc@{}}
\toprule
Test examples   & Model & 3D Stress Distribution & Stiffener webs and Flanges& Stress Range \\
\midrule
\multirow{2}{*}{1}  & GNN  & \includegraphics[width=0.25\textwidth]{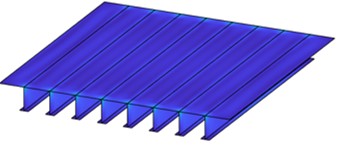} & \includegraphics[width=0.25\textwidth]{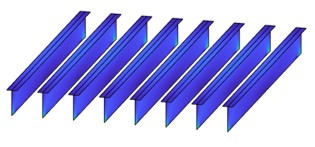} &\multirow{2}{*}{\includegraphics[width=0.05\textwidth]{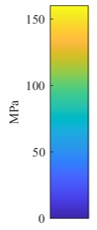}}\\
   & FEA  & \includegraphics[width=0.25\textwidth]{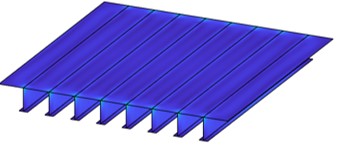}& \includegraphics[width=0.25\textwidth]{figures/Case1/Case1_14_3Dback_pred.jpg}& \\
\midrule
\multirow{2}{*}{2}    & GNN  &  \includegraphics[width=0.25\textwidth]{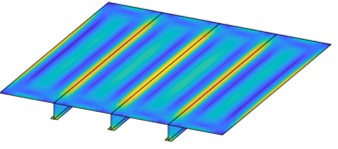} & \includegraphics[width=0.25\textwidth]{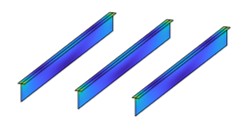} &\multirow{2}{*}{\includegraphics[width=0.05\textwidth]{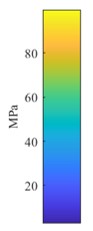}}\\
   & FEA  & \includegraphics[width=0.25\textwidth]{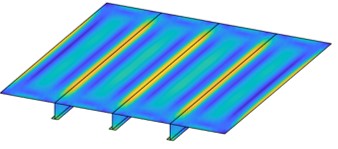}& \includegraphics[width=0.25\textwidth]{figures/Case1/Case1_34_3Dback_pred.jpg}& \\
\bottomrule
\end{tabular}
\end{table}

\begin{table}[ht]
\caption{Structural geometry details in parametric study for boundary conditions.}\label{tabl: Case 1 geometry}%
\begin{tabular}{@{}lccc@{}}
\toprule
Category & Test Example 1 &  Test Example 2 & Unit \\
\midrule
Plate thickness    & 14.51   & 12.58  & mm  \\
Stiffener web thickness    & 9.08   & 16.62  & mm  \\
Stiffener web height    &  308.37  & 223.98  & mm  \\
Flange thickness  & 16.83  & 15.61 & mm \\
Flange width   & 111.12 & 86.52 & mm \\
Number of stiffeners  & 8 & 3 & $-$ \\
Uniform pressure & 0.071 & 0.072 & MPa \\
BCs @ Plate edges & Simply support & Fixed & $-$ \\
BCs @ Stiffener web and Flange edges & Free & Simply support & $-$ \\
\botrule
\end{tabular}
\end{table}

\begin{figure}[htp]
    \centering
    \begin{minipage}[b]{0.48\textwidth}
        \includegraphics[width=\textwidth]{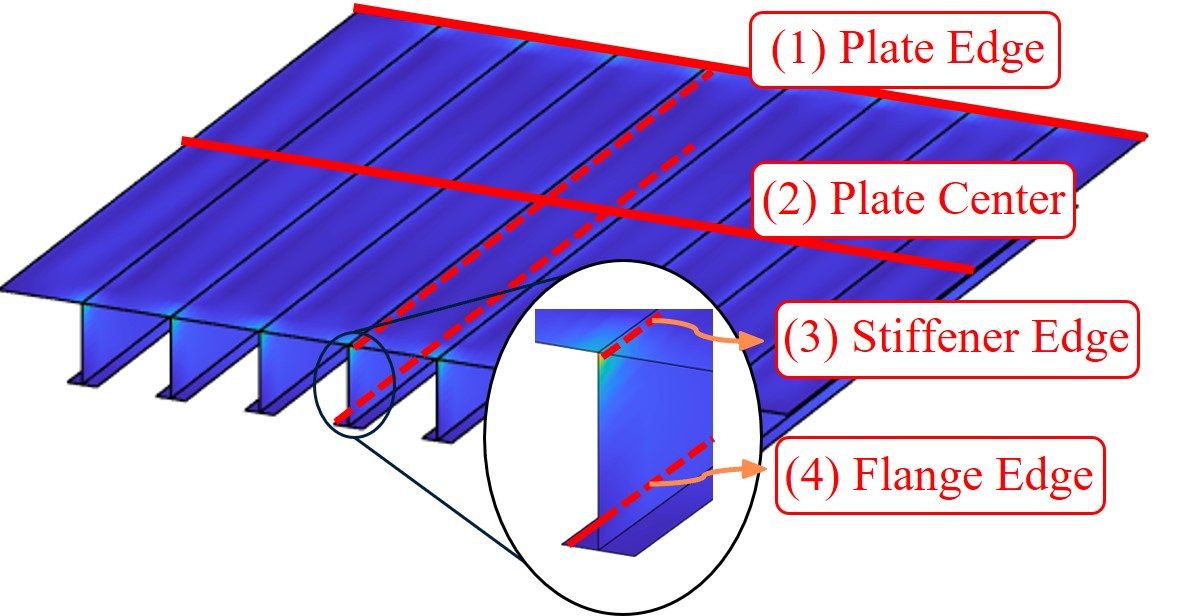}
        \\
        \subfloat[\normalsize Test example 1]{\includegraphics[width=\textwidth]{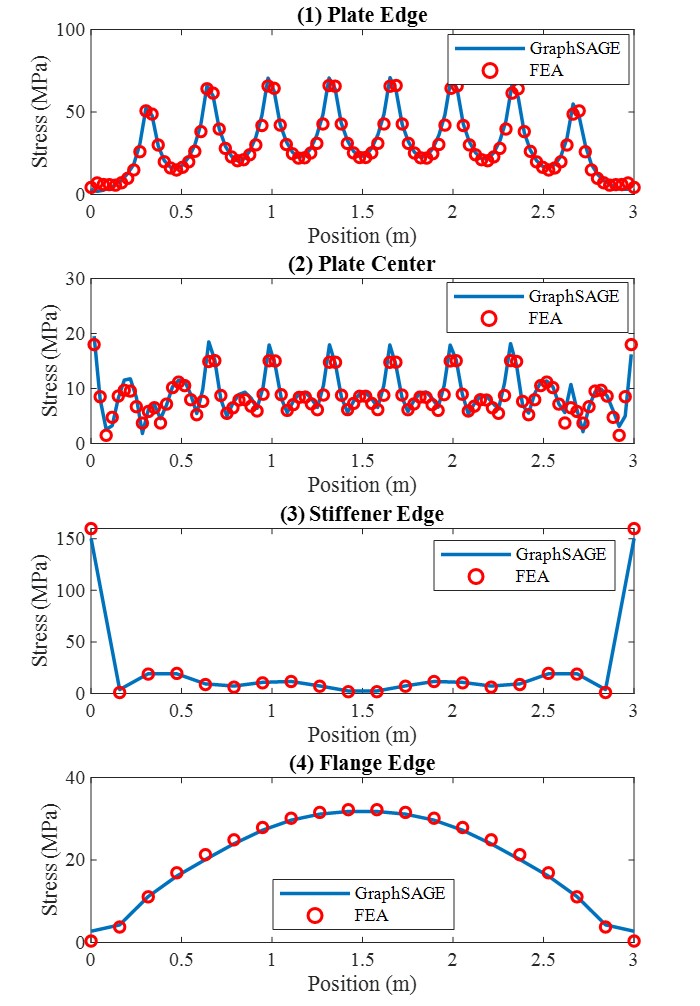}\label{fig: Case 1 plot (a)}}
    \end{minipage}
    \hfill
    \begin{minipage}[b]{0.48\textwidth}
        \includegraphics[width=\textwidth]{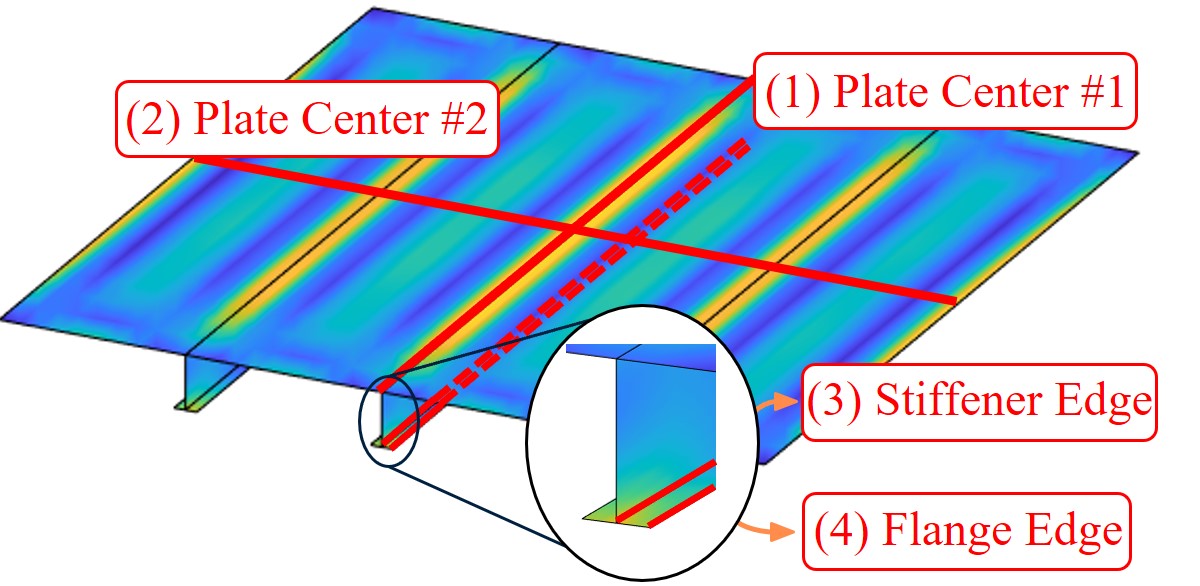}
        \\
        \subfloat[\normalsize Test example 2]{\includegraphics[width=\textwidth]{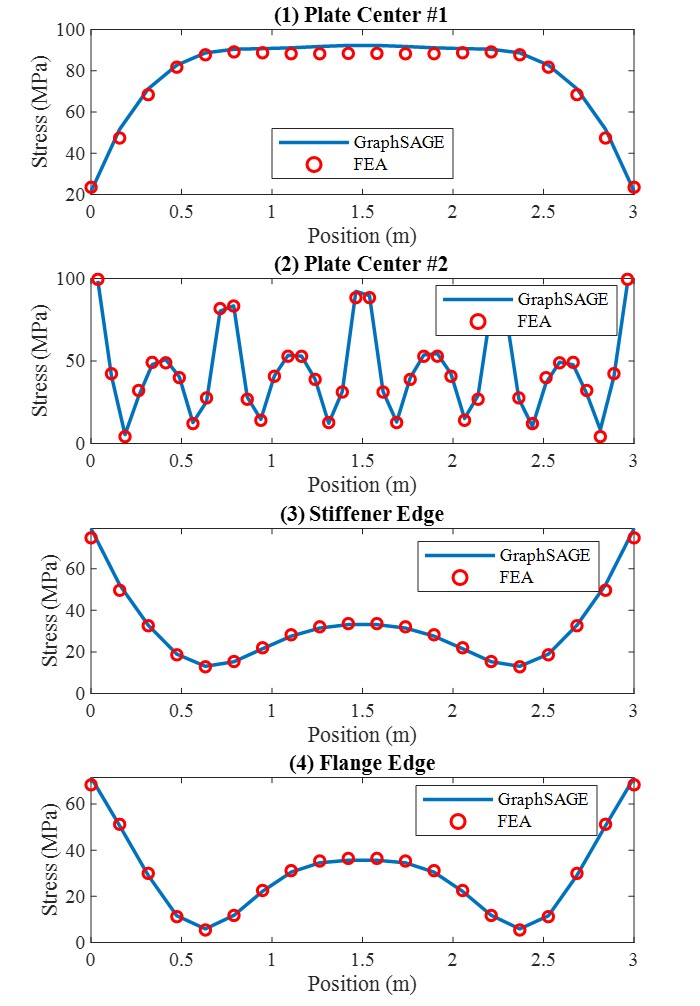}\label{fig: Case 1 plot (b)}}
    \end{minipage}
    \caption{Stress distribution comparison along specified paths in the stiffened Panel under parametric study with varying boundary conditions}\label{fig: Case 1 plot}
\end{figure}

\begin{figure}[htp]
\centering
\includegraphics[width=0.7\textwidth]{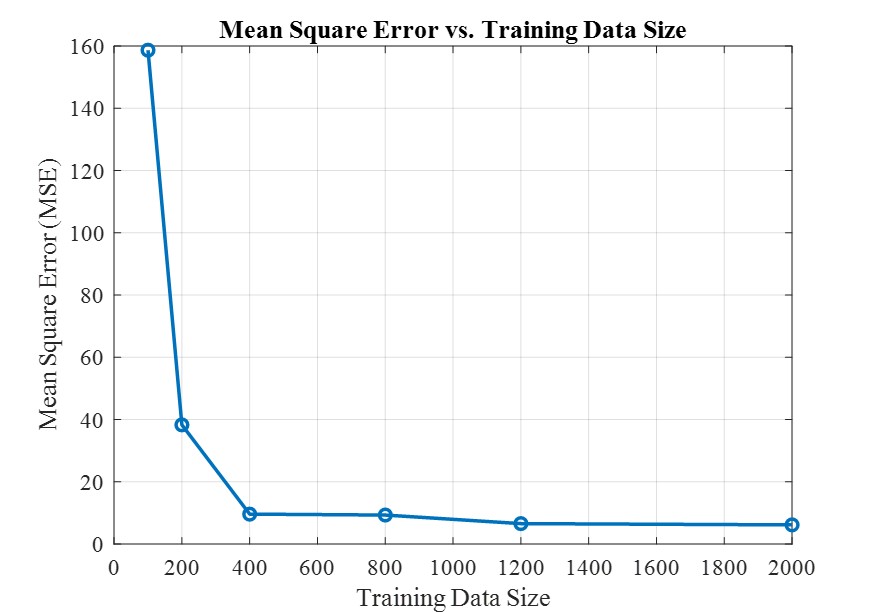}
\caption{GraphSAGE training data versus mean square error for parametric study considering the effect of boundary conditions.}\label{fig: Case 1 training data}
\end{figure}

In Table \ref{tabl: Case 1 contour}, we present a comparison between the predicted stress distribution of the stiffened panel and the distribution obtained through finite element analysis (FEA). We have selected two test examples at random to illustrate the prediction accuracy of our model. The 3D view of the structure is shown, where the contour shows the stress distribution and stress intensity accross the panel. For each test example, the same color for both GraphSAGE prediction and FEA ground truth represents the same stress value. The geometry of the structure and the external loading amplitude for both test examples can be found in Table \ref{tabl: Case 1 geometry}.

For both test examples in Table \ref{tabl: Case 1 contour}, the prediction of the stress distribution in the plate and stiffeners shows a high consistency with the FEA. This can also be verified by the detailed stress comparison in Fig. \ref{fig: Case 1 plot}, where we exhibit the stress distribution at different locations of the stiffened panel. The red circles denote the FEA results and the blue solid curve denotes the predicted stress distribution along the specified path with GNN. The location chosen to be evaluated depends on the geometry and stress distribution of the stiffened panel. The focus is on the areas that experience relatively higher stress, such as plate edges and stiffener edges. 

We can observe that although both test examples in this test case have employed different boundary conditions, the GraphSAGE model with the proposed graph embedding accurately captures the stress distribution of the structure. The predicted stress distribution exhibits a high consistency with the FEA results for both test examples at different locations. Although the prediction shows some slight deviation when predicting the lower stresses such as the plate center area in test example 1, the trained GraphSAGE network exhibits good performance in capturing the high-stress distribution. The maximum stress occurs at the intersection of plate and stiffener web edges in test example 1, and at the plate center in test example 2. For the high-stress estimation in both test examples (Fig. \ref{fig: Case 1 plot (a)} (3) and Fig. \ref{fig: Case 1 plot (b)} (1)), the prediction error is less than 1\%. 

As is generally known, the amount of data and data quality are paramount for accurate predictions with neural networks. Fig. \ref{fig: Case 1 training data} depicts the relationship between training data size and the MSE value for the validation set. The GraphSAGE model, utilized in this study, demonstrates consistent performance when the amount of data is 400 and above. This suggests that the proposed approach holds promise for practical applications in practical situations, as it demands a relatively small amount of data.

\subsubsection{The effect of structural geometry}\label{subsecGeometry}
An inherent advantage of graph neural networks (GNNs) over other types of neural networks lies in their adaptability to structural geometric changes. In this section, we aim to demonstrate the capacity of GNNs with the proposed graph embedding technique. Instead of employing uniformly distributed stiffeners, we introduce randomness to the distribution of stiffeners. Furthermore, we allow variations in the height of each stiffener separately and width of each flange separately, aiming at introducing more complexity to the structural geometries.

\begin{table}[ht]
\centering
\caption{Comparison of GraphSAGE predictions and FEA results for stiffened panel with complex geometry.}\label{tabl: Case 3 contour}%
\begin{tabular}{@{}ccccc@{}}
\toprule
Test examples   & Model & 3D Stress Distribution & Stiffeners and Flanges& Stress Range \\
\midrule
\multirow{2}{*}{1}  & GNN  & \includegraphics[width=0.25\textwidth]{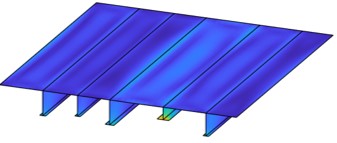} & \includegraphics[width=0.25\textwidth]{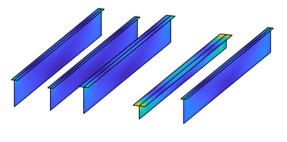} &\multirow{2}{*}{\includegraphics[width=0.05\textwidth]{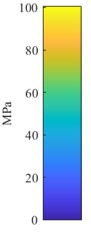}}\\
   & FEA  & \includegraphics[width=0.25\textwidth]{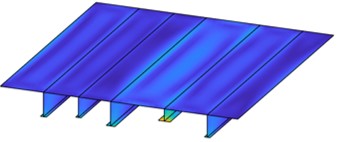}& \includegraphics[width=0.25\textwidth]{figures/Case3/Case3_187_3Dback_pred.jpg}& \\
\midrule
\multirow{2}{*}{2}    & GNN  & \includegraphics[width=0.25\textwidth]{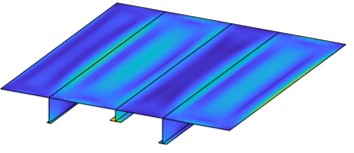} & \includegraphics[width=0.25\textwidth]{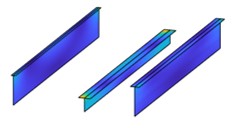} &\multirow{2}{*}{\includegraphics[width=0.05\textwidth]{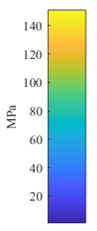}}\\
   & FEA  & \includegraphics[width=0.25\textwidth]{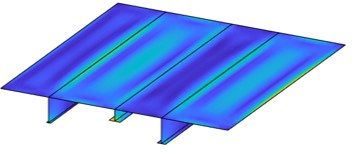}& \includegraphics[width=0.25\textwidth]{figures/Case3/Case3_46_3Dback_pred.jpg}& \\
\bottomrule
\end{tabular}
\end{table}

\begin{table}[ht]
\caption{Structural geometry details for the two test examples when considering complex geometry.}\label{tabl: Case 3 geometry}%
\begin{tabular}{@{}lccc@{}}
\toprule
Category & Test Example 1 &  Test Example 2 & Unit \\
\midrule
Plate thickness    &  19.85 & 16.14  & mm  \\
Stiffener web thickness    &  (11.8, 17.9, 12.8, 12.6, 10.8)  & (19.8, 17.0, 11.5) & mm  \\
Stiffener web height    & (320, 267, 320, 160, 293)   & (347,	160, 347) & mm  \\
Flange thickness  & (14.9, 19.0, 14.0, 6.7, 6.2) & (14.1, 14.8, 9.8) & mm \\
Flange width   &(74.3, 66.7, 107.8, 146.6, 69.8) & (52.1, 142.2, 104.0) & mm \\
Number of stiffeners  &5  & 3 & $-$ \\
Uniform pressure &0.079 & 0.090 & MPa \\
BCs @ Plate edges & Fixed & Fixed & $-$ \\
BCs @ Stiffener and Flange edges & Fixed & Fixed & $-$ \\
\botrule
\end{tabular}
\end{table}

\begin{figure}[htp]
    \centering
    \begin{minipage}[b]{0.48\textwidth}
        \includegraphics[width=\textwidth]{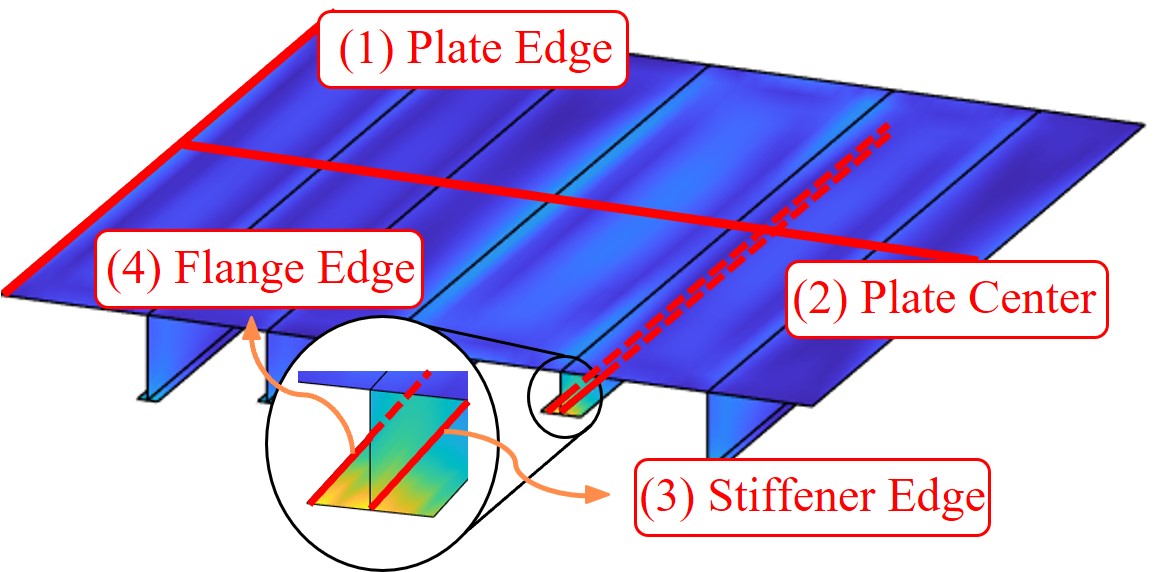}
        \\
        \subfloat[\normalsize Test example 1]{\includegraphics[width=\textwidth]{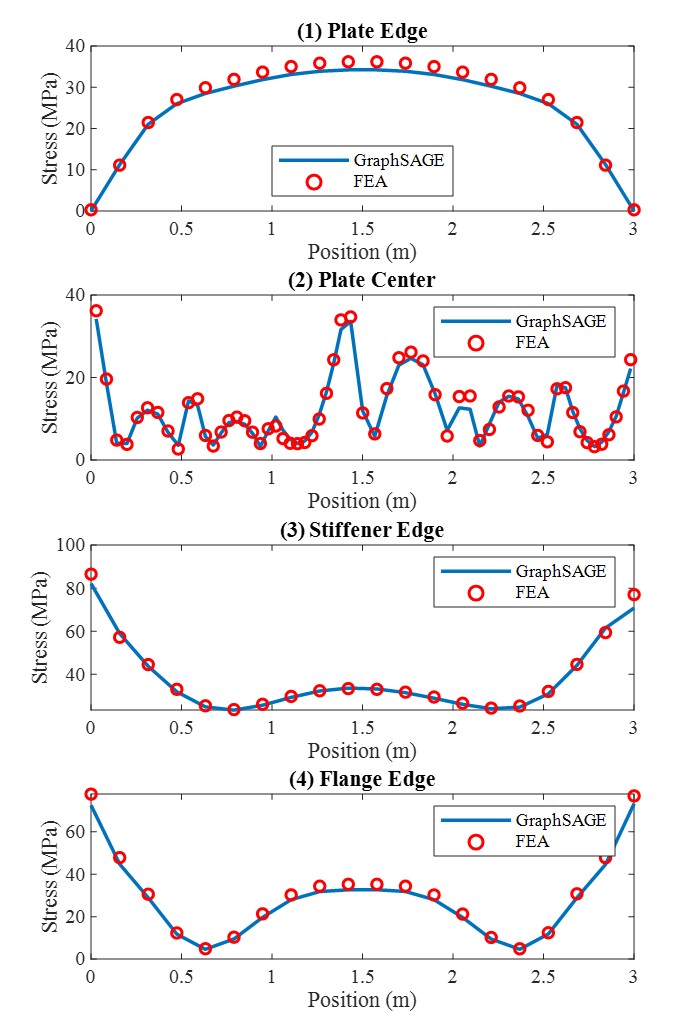}\label{fig: Case 3 plot(a)}}
    \end{minipage}
    \hfill
    \begin{minipage}[b]{0.48\textwidth}
        \includegraphics[width=\textwidth]{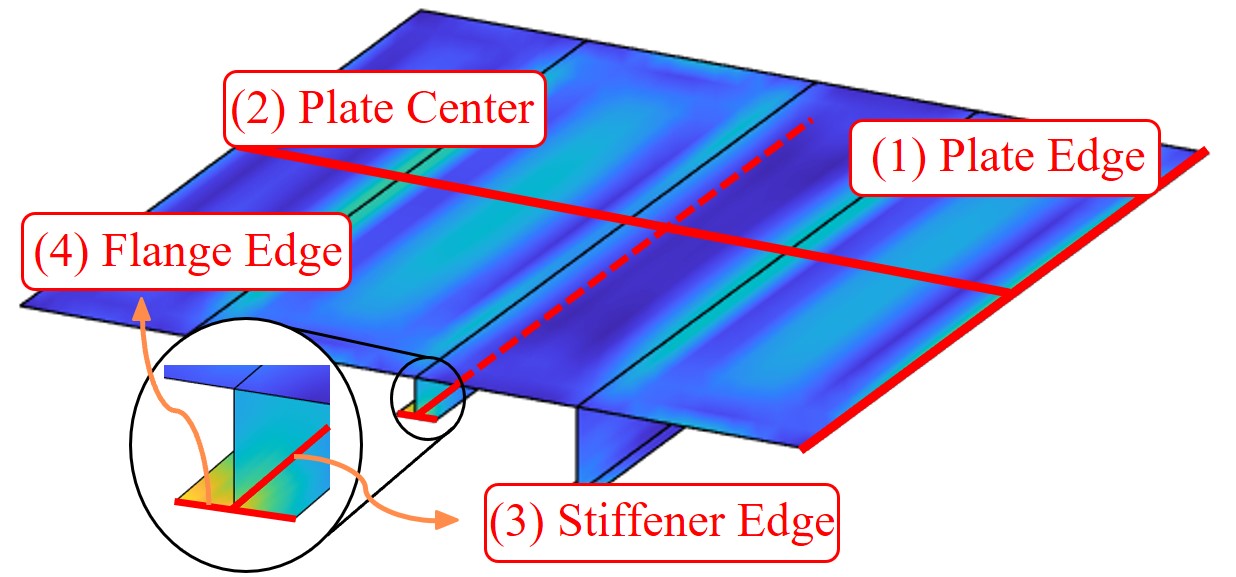}
        \\
        \subfloat[\normalsize Test example 2]{\includegraphics[width=\textwidth]{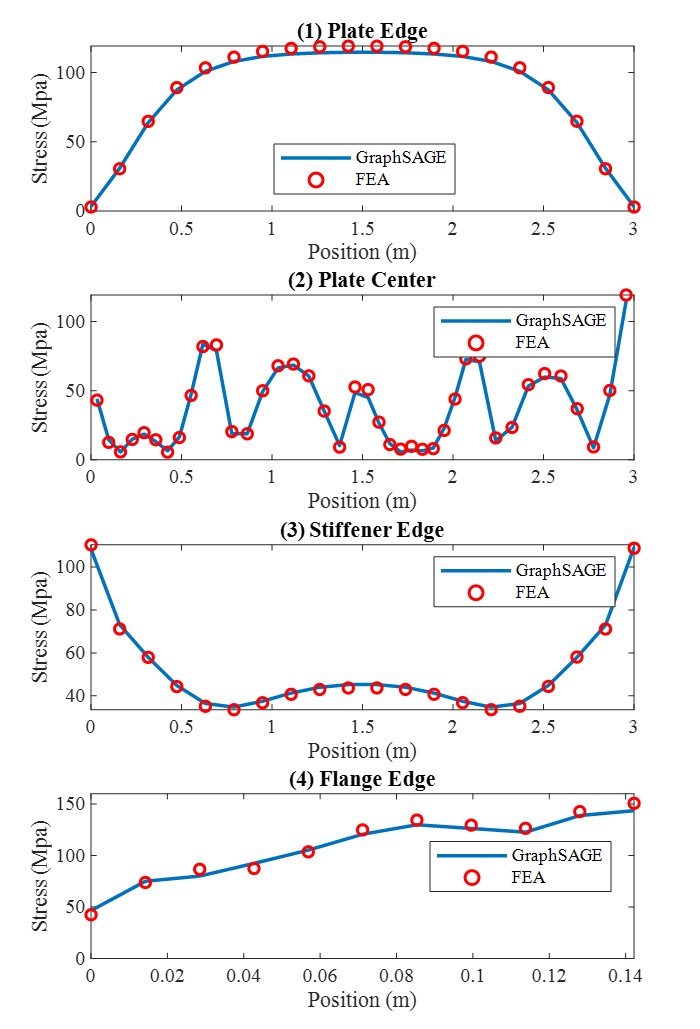}\label{fig: Case 3 plot(b)}}
    \end{minipage}
    \caption{Stress distribution comparison along specified paths in the stiffened panel under parametric study with complex structural geometry.}\label{fig: Case 3 plot}
\end{figure}

\begin{figure}[htp]
\centering
\includegraphics[width=0.7\textwidth]{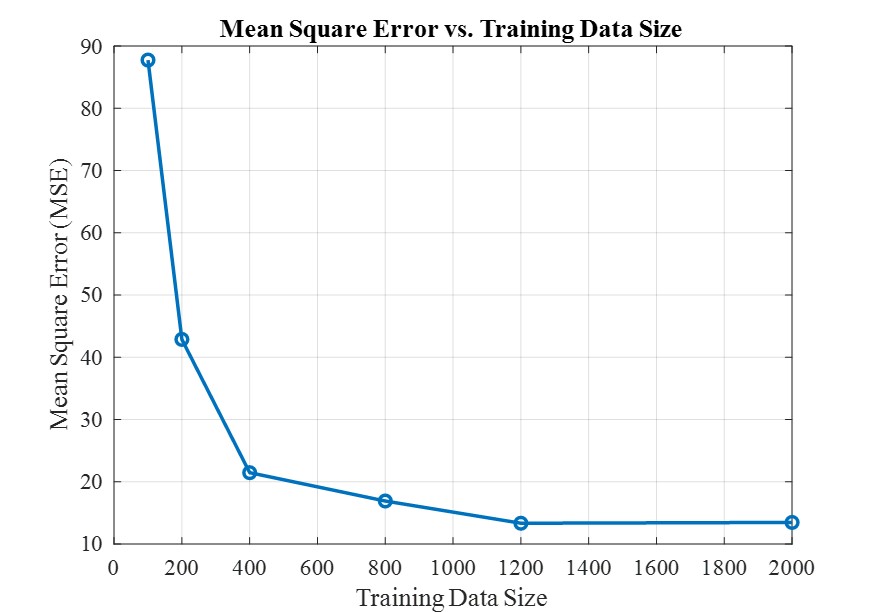}
\caption{GraphSAGE training data versus mean square error for parametric study with complex structural geometry}\label{fig: Case 3 training data}
\end{figure}

Specifically, in this section, the geometry of each stiffener and their corresponding flanges can be individually changed within the domain as set in Table \ref{tabl: var limit}, and they can be randomly positioned on the stiffened panel. With the remaining structural geometric parameters the same as in Section \ref{subsecBC}, two test examples are exhibited and their corresponding stress contour and detailed stress analysis are shown in Table \ref{tabl: Case 3 contour} and Fig. \ref{fig: Case 3 plot}, respectively. The corresponding structural geometry details can be found in Table. \ref{tabl: Case 3 geometry}. In each bracket, the numerical values represent the dimensions of the corresponding category of structural geometry, with the order of the numbers corresponding to the sequence of the stiffeners from left to right as shown in the Table. \ref{tabl: Case 3 contour}.

From Table \ref{tabl: Case 3 contour}, we can observe that increasing the complexity of the structural geometry does not hinder the prediction capability of the GraphSAGE model with the proposed graph embedding. For both test examples, the predicted stress contour for both plate and stiffeners exhibits a good consistency with the FEA results. As shown in Fig. \ref{fig: Case 3 plot(a)}, the predicted stress distribution shows a good alignment with the FEA results for all specified paths including the stiffener edge and flange edge, where the maximum stress occurs. The GraphSAGE estimation for the second test example is also accurate and captures all the stress information of the structure. The maximum prediction error occurs at the plate edge, which is about 6.18\%. The average prediction accuracy for the other paths is 96.7\%.

Following a similar methodology as discussed in Section \ref{subsecBC}, we examine the necessary amount of training data in this particular test case. In contrast to the previous scenario, the GraphSAGE model requires a larger quantity of training data to effectively capture stress distributions within more intricate structures. As depicted in Figure \ref{fig: Case 3 training data}, the curve reaches a plateau after having 1200 training data points. However, it is important to note that such a large dataset might not be required in practical real-life cases. The test structures considered in this subsection encompass lots of unrealistic structures, which can hinder the training process of the GraphSAGE model.

\section{Conclusion and future work}\label{sec5}

In this study, we explored the potential of graph neural networks (GNN), specifically Graph Sampling and Aggregation (GraphSAGE) as a promising avenue for developing a reduced-order model for stress prediction in stiffened panels. By proposing a novel graph embedding technique, the research showcased its superiority over conventional finite element-vertex embedding. A parametric analysis is conducted to demonstrate the versatility and robustness of the employed model, emphasizing its ability to handle various boundary conditions and complex geometric variations. The results indicate that the integration of GNN and the proposed graph embedding can revolutionize the structural modelling and analysis, offering both accuracy and efficiency. It can be concluded that:
\begin{itemize}
  \item The proposed structural (physical unit) entity-vertex graph embedding is a viable method to model the 3D structures, such as stiffened panels;
  \item The proposed structural entity-vertex embedding is more efficient than the conventional finite element-vertex embedding;
  \item GraphSAGE with the proposed embedding can handle various variables including boundary conditions and geometric variations;
  \item The predicted stress distribution exhibits high consistency with the FEA simulation.
\end{itemize}

Compared with the conventional reduced-order models employed in solid and structural mechanics, the proposed modelling technique in this study first introduces structural geometry as one of the input variables, which provides a foundation and potential for further development of reduced-order models for more complex structures. Even though this paper utilizes the data from the FEA simulation, obtaining the data in real-life is very difficult. However, this is a common issue for most data-driven models, which could be potentially addressed in future works as well.

\backmatter

\section*{Acknowledgments}

This research was supported by the Natural Sciences and Engineering Research Council of Canada (NSERC). This research was supported in part through computational resources and services provided by Advanced Research Computing at The University of British Columbia.

\bibliography{YCstiffenedPanel}


\begin{thebibliography}{44}
\ifx \bisbn   \undefined \def \bisbn  #1{ISBN #1}\fi
\ifx \binits  \undefined \def \binits#1{#1}\fi
\ifx \bauthor  \undefined \def \bauthor#1{#1}\fi
\ifx \batitle  \undefined \def \batitle#1{#1}\fi
\ifx \bjtitle  \undefined \def \bjtitle#1{#1}\fi
\ifx \bvolume  \undefined \def \bvolume#1{\textbf{#1}}\fi
\ifx \byear  \undefined \def \byear#1{#1}\fi
\ifx \bissue  \undefined \def \bissue#1{#1}\fi
\ifx \bfpage  \undefined \def \bfpage#1{#1}\fi
\ifx \blpage  \undefined \def \blpage #1{#1}\fi
\ifx \burl  \undefined \def \burl#1{\textsf{#1}}\fi
\ifx \doiurl  \undefined \def \doiurl#1{\url{https://doi.org/#1}}\fi
\ifx \betal  \undefined \def \betal{\textit{et al.}}\fi
\ifx \binstitute  \undefined \def \binstitute#1{#1}\fi
\ifx \binstitutionaled  \undefined \def \binstitutionaled#1{#1}\fi
\ifx \bctitle  \undefined \def \bctitle#1{#1}\fi
\ifx \beditor  \undefined \def \beditor#1{#1}\fi
\ifx \bpublisher  \undefined \def \bpublisher#1{#1}\fi
\ifx \bbtitle  \undefined \def \bbtitle#1{#1}\fi
\ifx \bedition  \undefined \def \bedition#1{#1}\fi
\ifx \bseriesno  \undefined \def \bseriesno#1{#1}\fi
\ifx \blocation  \undefined \def \blocation#1{#1}\fi
\ifx \bsertitle  \undefined \def \bsertitle#1{#1}\fi
\ifx \bsnm \undefined \def \bsnm#1{#1}\fi
\ifx \bsuffix \undefined \def \bsuffix#1{#1}\fi
\ifx \bparticle \undefined \def \bparticle#1{#1}\fi
\ifx \barticle \undefined \def \barticle#1{#1}\fi
\bibcommenthead
\ifx \bconfdate \undefined \def \bconfdate #1{#1}\fi
\ifx \botherref \undefined \def \botherref #1{#1}\fi
\ifx \url \undefined \def \url#1{\textsf{#1}}\fi
\ifx \bchapter \undefined \def \bchapter#1{#1}\fi
\ifx \bbook \undefined \def \bbook#1{#1}\fi
\ifx \bcomment \undefined \def \bcomment#1{#1}\fi
\ifx \oauthor \undefined \def \oauthor#1{#1}\fi
\ifx \citeauthoryear \undefined \def \citeauthoryear#1{#1}\fi
\ifx \endbibitem  \undefined \def \endbibitem {}\fi
\ifx \bconflocation  \undefined \def \bconflocation#1{#1}\fi
\ifx \arxivurl  \undefined \def \arxivurl#1{\textsf{#1}}\fi
\csname PreBibitemsHook\endcsname

\bibitem[\protect\citeauthoryear{Clough}{1990}]{clough1990original}
\begin{barticle}
\bauthor{\bsnm{Clough}, \binits{R.W.}}:
\batitle{Original formulation of the finite element method}.
\bjtitle{Finite elements in analysis and design}
\bvolume{7}(\bissue{2}),
\bfpage{89}--\blpage{101}
(\byear{1990})
\end{barticle}
\endbibitem

\bibitem[\protect\citeauthoryear{Sullivan et~al.}{2018}]{sullivan2018effect}
\begin{barticle}
\bauthor{\bsnm{Sullivan}, \binits{J.L.}},
\bauthor{\bsnm{Lewis}, \binits{G.M.}},
\bauthor{\bsnm{Keoleian}, \binits{G.A.}}:
\batitle{Effect of mass on multimodal fuel consumption in moving people and
  freight in the us}.
\bjtitle{Transportation Research Part D: Transport and Environment}
\bvolume{63},
\bfpage{786}--\blpage{808}
(\byear{2018})
\end{barticle}
\endbibitem

\bibitem[\protect\citeauthoryear{Samanipour and
  Jelovica}{2020}]{samanipour2020adaptive}
\begin{barticle}
\bauthor{\bsnm{Samanipour}, \binits{F.}},
\bauthor{\bsnm{Jelovica}, \binits{J.}}:
\batitle{Adaptive repair method for constraint handling in multi-objective
  genetic algorithm based on relationship between constraints and variables}.
\bjtitle{Applied Soft Computing}
\bvolume{90},
\bfpage{106143}
(\byear{2020})
\end{barticle}
\endbibitem

\bibitem[\protect\citeauthoryear{Jelovica and Cai}{2022}]{jelovica2022improved}
\begin{botherref}
\oauthor{\bsnm{Jelovica}, \binits{J.}},
\oauthor{\bsnm{Cai}, \binits{Y.}}:
Improved multi-objective structural optimization with adaptive repair-based
  constraint handling.
Engineering Optimization,
1--20
(2022)
\end{botherref}
\endbibitem

\bibitem[\protect\citeauthoryear{Chu et~al.}{2021}]{chu2021design}
\begin{barticle}
\bauthor{\bsnm{Chu}, \binits{S.}},
\bauthor{\bsnm{Featherston}, \binits{C.}},
\bauthor{\bsnm{Kim}, \binits{H.A.}}:
\batitle{Design of stiffened panels for stress and buckling via topology
  optimization}.
\bjtitle{Structural and Multidisciplinary Optimization}
\bvolume{64},
\bfpage{3123}--\blpage{3146}
(\byear{2021})
\end{barticle}
\endbibitem

\bibitem[\protect\citeauthoryear{Chen et~al.}{2006}]{chen2006review}
\begin{barticle}
\bauthor{\bsnm{Chen}, \binits{V.C.}},
\bauthor{\bsnm{Tsui}, \binits{K.-L.}},
\bauthor{\bsnm{Barton}, \binits{R.R.}},
\bauthor{\bsnm{Meckesheimer}, \binits{M.}}:
\batitle{A review on design, modeling and applications of computer
  experiments}.
\bjtitle{IIE transactions}
\bvolume{38}(\bissue{4}),
\bfpage{273}--\blpage{291}
(\byear{2006})
\end{barticle}
\endbibitem

\bibitem[\protect\citeauthoryear{Mai et~al.}{2022}]{mai2022robust}
\begin{barticle}
\bauthor{\bsnm{Mai}, \binits{H.T.}},
\bauthor{\bsnm{Lieu}, \binits{Q.X.}},
\bauthor{\bsnm{Kang}, \binits{J.}},
\bauthor{\bsnm{Lee}, \binits{J.}}:
\batitle{A robust unsupervised neural network framework for geometrically
  nonlinear analysis of inelastic truss structures}.
\bjtitle{Applied Mathematical Modelling}
\bvolume{107},
\bfpage{332}--\blpage{352}
(\byear{2022})
\end{barticle}
\endbibitem

\bibitem[\protect\citeauthoryear{Shojaeefard
  et~al.}{2013}]{shojaeefard2013modelling}
\begin{barticle}
\bauthor{\bsnm{Shojaeefard}, \binits{M.H.}},
\bauthor{\bsnm{Behnagh}, \binits{R.A.}},
\bauthor{\bsnm{Akbari}, \binits{M.}},
\bauthor{\bsnm{Givi}, \binits{M.K.B.}},
\bauthor{\bsnm{Farhani}, \binits{F.}}:
\batitle{Modelling and pareto optimization of mechanical properties of friction
  stir welded aa7075/aa5083 butt joints using neural network and particle swarm
  algorithm}.
\bjtitle{Materials \& Design}
\bvolume{44},
\bfpage{190}--\blpage{198}
(\byear{2013})
\end{barticle}
\endbibitem

\bibitem[\protect\citeauthoryear{Kabir et~al.}{2021}]{kabir2021failure}
\begin{barticle}
\bauthor{\bsnm{Kabir}, \binits{M.A.B.}},
\bauthor{\bsnm{Hasan}, \binits{A.S.}},
\bauthor{\bsnm{Billah}, \binits{A.M.}}:
\batitle{Failure mode identification of column base plate connection using
  data-driven machine learning techniques}.
\bjtitle{Engineering Structures}
\bvolume{240},
\bfpage{112389}
(\byear{2021})
\end{barticle}
\endbibitem

\bibitem[\protect\citeauthoryear{Hornik et~al.}{1989}]{hornik1989multilayer}
\begin{barticle}
\bauthor{\bsnm{Hornik}, \binits{K.}},
\bauthor{\bsnm{Stinchcombe}, \binits{M.}},
\bauthor{\bsnm{White}, \binits{H.}}:
\batitle{Multilayer feedforward networks are universal approximators}.
\bjtitle{Neural networks}
\bvolume{2}(\bissue{5}),
\bfpage{359}--\blpage{366}
(\byear{1989})
\end{barticle}
\endbibitem

\bibitem[\protect\citeauthoryear{Papadrakakis
  et~al.}{1998}]{papadrakakis1998structural}
\begin{barticle}
\bauthor{\bsnm{Papadrakakis}, \binits{M.}},
\bauthor{\bsnm{Lagaros}, \binits{N.D.}},
\bauthor{\bsnm{Tsompanakis}, \binits{Y.}}:
\batitle{Structural optimization using evolution strategies and neural
  networks}.
\bjtitle{Computer methods in applied mechanics and engineering}
\bvolume{156}(\bissue{1-4}),
\bfpage{309}--\blpage{333}
(\byear{1998})
\end{barticle}
\endbibitem

\bibitem[\protect\citeauthoryear{Bisagni and Lanzi}{2002}]{bisagni2002post}
\begin{barticle}
\bauthor{\bsnm{Bisagni}, \binits{C.}},
\bauthor{\bsnm{Lanzi}, \binits{L.}}:
\batitle{Post-buckling optimisation of composite stiffened panels using neural
  networks}.
\bjtitle{Composite Structures}
\bvolume{58}(\bissue{2}),
\bfpage{237}--\blpage{247}
(\byear{2002})
\end{barticle}
\endbibitem

\bibitem[\protect\citeauthoryear{Sun et~al.}{2021}]{sun2021prediction}
\begin{barticle}
\bauthor{\bsnm{Sun}, \binits{Z.}},
\bauthor{\bsnm{Lei}, \binits{Z.}},
\bauthor{\bsnm{Bai}, \binits{R.}},
\bauthor{\bsnm{Jiang}, \binits{H.}},
\bauthor{\bsnm{Zou}, \binits{J.}},
\bauthor{\bsnm{Ma}, \binits{Y.}},
\bauthor{\bsnm{Yan}, \binits{C.}}:
\batitle{Prediction of compression buckling load and buckling mode of
  hat-stiffened panels using artificial neural network}.
\bjtitle{Engineering Structures}
\bvolume{242},
\bfpage{112275}
(\byear{2021})
\end{barticle}
\endbibitem

\bibitem[\protect\citeauthoryear{Limbachiya and
  Shamass}{2021}]{limbachiya2021application}
\begin{barticle}
\bauthor{\bsnm{Limbachiya}, \binits{V.}},
\bauthor{\bsnm{Shamass}, \binits{R.}}:
\batitle{Application of artificial neural networks for web-post shear
  resistance of cellular steel beams}.
\bjtitle{Thin-Walled Structures}
\bvolume{161},
\bfpage{107414}
(\byear{2021})
\end{barticle}
\endbibitem

\bibitem[\protect\citeauthoryear{Ferreira et~al.}{2022}]{ferreira2022lateral}
\begin{barticle}
\bauthor{\bsnm{Ferreira}, \binits{F.P.V.}},
\bauthor{\bsnm{Shamass}, \binits{R.}},
\bauthor{\bsnm{Limbachiya}, \binits{V.}},
\bauthor{\bsnm{Tsavdaridis}, \binits{K.D.}},
\bauthor{\bsnm{Martins}, \binits{C.H.}}:
\batitle{Lateral--torsional buckling resistance prediction model for steel
  cellular beams generated by artificial neural networks (ann)}.
\bjtitle{Thin-Walled Structures}
\bvolume{170},
\bfpage{108592}
(\byear{2022})
\end{barticle}
\endbibitem

\bibitem[\protect\citeauthoryear{Mandal et~al.}{2021}]{mandal2021application}
\begin{barticle}
\bauthor{\bsnm{Mandal}, \binits{P.}},
\bauthor{\bsnm{Adil}, \binits{M.T.}},
\bauthor{\bsnm{Naz}, \binits{F.}}, \betal:
\batitle{Application of artificial neural network to predict buckling load of
  thin cylindrical shells under axial compression}.
\bjtitle{Engineering Structures}
\bvolume{248},
\bfpage{113221}
(\byear{2021})
\end{barticle}
\endbibitem

\bibitem[\protect\citeauthoryear{Zarringol
  et~al.}{2023}]{zarringol2023artificial}
\begin{barticle}
\bauthor{\bsnm{Zarringol}, \binits{M.}},
\bauthor{\bsnm{Patel}, \binits{V.I.}},
\bauthor{\bsnm{Liang}, \binits{Q.Q.}}:
\batitle{Artificial neural network model for strength predictions of cfst
  columns strengthened with cfrp}.
\bjtitle{Engineering Structures}
\bvolume{281},
\bfpage{115784}
(\byear{2023})
\end{barticle}
\endbibitem

\bibitem[\protect\citeauthoryear{Zhu et~al.}{2023}]{zhu2023artificial}
\begin{barticle}
\bauthor{\bsnm{Zhu}, \binits{M.}},
\bauthor{\bsnm{Peng}, \binits{Y.}},
\bauthor{\bsnm{Ma}, \binits{W.}},
\bauthor{\bsnm{Guo}, \binits{J.}},
\bauthor{\bsnm{Lu}, \binits{J.}}:
\batitle{Artificial neural network-aided force finding of cable dome structures
  with diverse integral self-stress states-framework and case study}.
\bjtitle{Engineering Structures}
\bvolume{285},
\bfpage{116004}
(\byear{2023})
\end{barticle}
\endbibitem

\bibitem[\protect\citeauthoryear{Ramkumar
  et~al.}{2021}]{ramkumar2021unconventional}
\begin{barticle}
\bauthor{\bsnm{Ramkumar}, \binits{G.}},
\bauthor{\bsnm{Sahoo}, \binits{S.}},
\bauthor{\bsnm{Anitha}, \binits{G.}},
\bauthor{\bsnm{Ramesh}, \binits{S.}},
\bauthor{\bsnm{Nirmala}, \binits{P.}},
\bauthor{\bsnm{Tamilselvi}, \binits{M.}},
\bauthor{\bsnm{Subbiah}, \binits{R.}},
\bauthor{\bsnm{Rajkumar}, \binits{S.}}:
\batitle{An unconventional approach for analyzing the mechanical properties of
  natural fiber composite using convolutional neural network}.
\bjtitle{Advances in Materials Science and Engineering}
\bvolume{2021},
\bfpage{1}--\blpage{15}
(\byear{2021})
\end{barticle}
\endbibitem

\bibitem[\protect\citeauthoryear{Banga et~al.}{2018}]{banga20183d}
\begin{botherref}
\oauthor{\bsnm{Banga}, \binits{S.}},
\oauthor{\bsnm{Gehani}, \binits{H.}},
\oauthor{\bsnm{Bhilare}, \binits{S.}},
\oauthor{\bsnm{Patel}, \binits{S.}},
\oauthor{\bsnm{Kara}, \binits{L.}}:
3d topology optimization using convolutional neural networks.
arXiv preprint arXiv:1808.07440
(2018)
\end{botherref}
\endbibitem

\bibitem[\protect\citeauthoryear{Wang et~al.}{2021}]{wang2021stressnet}
\begin{barticle}
\bauthor{\bsnm{Wang}, \binits{Y.}},
\bauthor{\bsnm{Oyen}, \binits{D.}},
\bauthor{\bsnm{Guo}, \binits{W.}},
\bauthor{\bsnm{Mehta}, \binits{A.}},
\bauthor{\bsnm{Scott}, \binits{C.B.}},
\bauthor{\bsnm{Panda}, \binits{N.}},
\bauthor{\bsnm{Fern{\'a}ndez-Godino}, \binits{M.G.}},
\bauthor{\bsnm{Srinivasan}, \binits{G.}},
\bauthor{\bsnm{Yue}, \binits{X.}}:
\batitle{Stressnet-deep learning to predict stress with fracture propagation in
  brittle materials}.
\bjtitle{npj Materials Degradation}
\bvolume{5}(\bissue{1}),
\bfpage{6}
(\byear{2021})
\end{barticle}
\endbibitem

\bibitem[\protect\citeauthoryear{Bolandi et~al.}{2022}]{bolandi2022deep}
\begin{barticle}
\bauthor{\bsnm{Bolandi}, \binits{H.}},
\bauthor{\bsnm{Li}, \binits{X.}},
\bauthor{\bsnm{Salem}, \binits{T.}},
\bauthor{\bsnm{Boddeti}, \binits{V.N.}},
\bauthor{\bsnm{Lajnef}, \binits{N.}}:
\batitle{Deep learning paradigm for prediction of stress distribution in
  damaged structural components with stress concentrations}.
\bjtitle{Advances in Engineering Software}
\bvolume{173},
\bfpage{103240}
(\byear{2022})
\end{barticle}
\endbibitem

\bibitem[\protect\citeauthoryear{Nie et~al.}{2021}]{nie2021topologygan}
\begin{barticle}
\bauthor{\bsnm{Nie}, \binits{Z.}},
\bauthor{\bsnm{Lin}, \binits{T.}},
\bauthor{\bsnm{Jiang}, \binits{H.}},
\bauthor{\bsnm{Kara}, \binits{L.B.}}:
\batitle{Topologygan: Topology optimization using generative adversarial
  networks based on physical fields over the initial domain}.
\bjtitle{Journal of Mechanical Design}
\bvolume{143}(\bissue{3}),
\bfpage{031715}
(\byear{2021})
\end{barticle}
\endbibitem

\bibitem[\protect\citeauthoryear{Achour et~al.}{2020}]{achour2020development}
\begin{bchapter}
\bauthor{\bsnm{Achour}, \binits{G.}},
\bauthor{\bsnm{Sung}, \binits{W.J.}},
\bauthor{\bsnm{Pinon-Fischer}, \binits{O.J.}},
\bauthor{\bsnm{Mavris}, \binits{D.N.}}:
\bctitle{Development of a conditional generative adversarial network for
  airfoil shape optimization}.
In: \bbtitle{AIAA Scitech 2020 Forum},
p. \bfpage{2261}
(\byear{2020})
\end{bchapter}
\endbibitem

\bibitem[\protect\citeauthoryear{Mao et~al.}{2020}]{mao2020designing}
\begin{barticle}
\bauthor{\bsnm{Mao}, \binits{Y.}},
\bauthor{\bsnm{He}, \binits{Q.}},
\bauthor{\bsnm{Zhao}, \binits{X.}}:
\batitle{Designing complex architectured materials with generative adversarial
  networks}.
\bjtitle{Science advances}
\bvolume{6}(\bissue{17}),
\bfpage{4169}
(\byear{2020})
\end{barticle}
\endbibitem

\bibitem[\protect\citeauthoryear{Shu et~al.}{2020}]{shu20203d}
\begin{barticle}
\bauthor{\bsnm{Shu}, \binits{D.}},
\bauthor{\bsnm{Cunningham}, \binits{J.}},
\bauthor{\bsnm{Stump}, \binits{G.}},
\bauthor{\bsnm{Miller}, \binits{S.W.}},
\bauthor{\bsnm{Yukish}, \binits{M.A.}},
\bauthor{\bsnm{Simpson}, \binits{T.W.}},
\bauthor{\bsnm{Tucker}, \binits{C.S.}}:
\batitle{3d design using generative adversarial networks and physics-based
  validation}.
\bjtitle{Journal of Mechanical Design}
\bvolume{142}(\bissue{7}),
\bfpage{071701}
(\byear{2020})
\end{barticle}
\endbibitem

\bibitem[\protect\citeauthoryear{Xu et~al.}{2017}]{xu2017scene}
\begin{bchapter}
\bauthor{\bsnm{Xu}, \binits{D.}},
\bauthor{\bsnm{Zhu}, \binits{Y.}},
\bauthor{\bsnm{Choy}, \binits{C.B.}},
\bauthor{\bsnm{Fei-Fei}, \binits{L.}}:
\bctitle{Scene graph generation by iterative message passing}.
In: \bbtitle{Proceedings of the IEEE Conference on Computer Vision and Pattern
  Recognition},
pp. \bfpage{5410}--\blpage{5419}
(\byear{2017})
\end{bchapter}
\endbibitem

\bibitem[\protect\citeauthoryear{Yao et~al.}{2018}]{yao2018deep}
\begin{bchapter}
\bauthor{\bsnm{Yao}, \binits{H.}},
\bauthor{\bsnm{Wu}, \binits{F.}},
\bauthor{\bsnm{Ke}, \binits{J.}},
\bauthor{\bsnm{Tang}, \binits{X.}},
\bauthor{\bsnm{Jia}, \binits{Y.}},
\bauthor{\bsnm{Lu}, \binits{S.}},
\bauthor{\bsnm{Gong}, \binits{P.}},
\bauthor{\bsnm{Ye}, \binits{J.}},
\bauthor{\bsnm{Li}, \binits{Z.}}:
\bctitle{Deep multi-view spatial-temporal network for taxi demand prediction}.
In: \bbtitle{Proceedings of the AAAI Conference on Artificial Intelligence},
vol. \bseriesno{32}
(\byear{2018})
\end{bchapter}
\endbibitem

\bibitem[\protect\citeauthoryear{Gilmer et~al.}{2017}]{gilmer2017neural}
\begin{bchapter}
\bauthor{\bsnm{Gilmer}, \binits{J.}},
\bauthor{\bsnm{Schoenholz}, \binits{S.S.}},
\bauthor{\bsnm{Riley}, \binits{P.F.}},
\bauthor{\bsnm{Vinyals}, \binits{O.}},
\bauthor{\bsnm{Dahl}, \binits{G.E.}}:
\bctitle{Neural message passing for quantum chemistry}.
In: \bbtitle{International Conference on Machine Learning},
pp. \bfpage{1263}--\blpage{1272}
(\byear{2017}).
\bcomment{PMLR}
\end{bchapter}
\endbibitem

\bibitem[\protect\citeauthoryear{Fout et~al.}{2017}]{fout2017protein}
\begin{botherref}
\oauthor{\bsnm{Fout}, \binits{A.}},
\oauthor{\bsnm{Byrd}, \binits{J.}},
\oauthor{\bsnm{Shariat}, \binits{B.}},
\oauthor{\bsnm{Ben-Hur}, \binits{A.}}:
Protein interface prediction using graph convolutional networks.
Advances in neural information processing systems
\textbf{30}
(2017)
\end{botherref}
\endbibitem

\bibitem[\protect\citeauthoryear{Ying et~al.}{2018}]{ying2018graph}
\begin{bchapter}
\bauthor{\bsnm{Ying}, \binits{R.}},
\bauthor{\bsnm{He}, \binits{R.}},
\bauthor{\bsnm{Chen}, \binits{K.}},
\bauthor{\bsnm{Eksombatchai}, \binits{P.}},
\bauthor{\bsnm{Hamilton}, \binits{W.L.}},
\bauthor{\bsnm{Leskovec}, \binits{J.}}:
\bctitle{Graph convolutional neural networks for web-scale recommender
  systems}.
In: \bbtitle{Proceedings of the 24th ACM SIGKDD International Conference on
  Knowledge Discovery \& Data Mining},
pp. \bfpage{974}--\blpage{983}
(\byear{2018})
\end{bchapter}
\endbibitem

\bibitem[\protect\citeauthoryear{Lino et~al.}{2022}]{lino2022multi}
\begin{botherref}
\oauthor{\bsnm{Lino}, \binits{M.}},
\oauthor{\bsnm{Fotiadis}, \binits{S.}},
\oauthor{\bsnm{Bharath}, \binits{A.A.}},
\oauthor{\bsnm{Cantwell}, \binits{C.D.}}:
Multi-scale rotation-equivariant graph neural networks for unsteady eulerian
  fluid dynamics.
Physics of Fluids
\textbf{34}(8)
(2022)
\end{botherref}
\endbibitem

\bibitem[\protect\citeauthoryear{Shao et~al.}{2023}]{shao2023pignn}
\begin{barticle}
\bauthor{\bsnm{Shao}, \binits{X.}},
\bauthor{\bsnm{Liu}, \binits{Z.}},
\bauthor{\bsnm{Zhang}, \binits{S.}},
\bauthor{\bsnm{Zhao}, \binits{Z.}},
\bauthor{\bsnm{Hu}, \binits{C.}}:
\batitle{Pignn-cfd: A physics-informed graph neural network for rapid
  predicting urban wind field defined on unstructured mesh}.
\bjtitle{Building and Environment}
\bvolume{232},
\bfpage{110056}
(\byear{2023})
\end{barticle}
\endbibitem

\bibitem[\protect\citeauthoryear{Pfaff et~al.}{2020}]{pfaff2020learning}
\begin{botherref}
\oauthor{\bsnm{Pfaff}, \binits{T.}},
\oauthor{\bsnm{Fortunato}, \binits{M.}},
\oauthor{\bsnm{Sanchez-Gonzalez}, \binits{A.}},
\oauthor{\bsnm{Battaglia}, \binits{P.W.}}:
Learning mesh-based simulation with graph networks.
arXiv preprint arXiv:2010.03409
(2020)
\end{botherref}
\endbibitem

\bibitem[\protect\citeauthoryear{Zheng et~al.}{2023}]{zheng2023tso}
\begin{barticle}
\bauthor{\bsnm{Zheng}, \binits{S.}},
\bauthor{\bsnm{Qiu}, \binits{L.}},
\bauthor{\bsnm{Lan}, \binits{F.}}:
\batitle{Tso-gcn: A graph convolutional network approach for real-time and
  generalizable truss structural optimization}.
\bjtitle{Applied Soft Computing}
\bvolume{134},
\bfpage{110015}
(\byear{2023})
\end{barticle}
\endbibitem

\bibitem[\protect\citeauthoryear{Bacciu et~al.}{2020}]{bacciu2020gentle}
\begin{barticle}
\bauthor{\bsnm{Bacciu}, \binits{D.}},
\bauthor{\bsnm{Errica}, \binits{F.}},
\bauthor{\bsnm{Micheli}, \binits{A.}},
\bauthor{\bsnm{Podda}, \binits{M.}}:
\batitle{A gentle introduction to deep learning for graphs}.
\bjtitle{Neural Networks}
\bvolume{129},
\bfpage{203}--\blpage{221}
(\byear{2020})
\end{barticle}
\endbibitem

\bibitem[\protect\citeauthoryear{Whalen and Mueller}{2022}]{whalen2022toward}
\begin{barticle}
\bauthor{\bsnm{Whalen}, \binits{E.}},
\bauthor{\bsnm{Mueller}, \binits{C.}}:
\batitle{Toward reusable surrogate models: Graph-based transfer learning on
  trusses}.
\bjtitle{Journal of Mechanical Design}
\bvolume{144}(\bissue{2}),
\bfpage{021704}
(\byear{2022})
\end{barticle}
\endbibitem

\bibitem[\protect\citeauthoryear{Bruna et~al.}{2013}]{bruna2013spectral}
\begin{botherref}
\oauthor{\bsnm{Bruna}, \binits{J.}},
\oauthor{\bsnm{Zaremba}, \binits{W.}},
\oauthor{\bsnm{Szlam}, \binits{A.}},
\oauthor{\bsnm{LeCun}, \binits{Y.}}:
Spectral networks and locally connected networks on graphs.
arXiv preprint arXiv:1312.6203
(2013)
\end{botherref}
\endbibitem

\bibitem[\protect\citeauthoryear{Defferrard
  et~al.}{2016}]{defferrard2016convolutional}
\begin{botherref}
\oauthor{\bsnm{Defferrard}, \binits{M.}},
\oauthor{\bsnm{Bresson}, \binits{X.}},
\oauthor{\bsnm{Vandergheynst}, \binits{P.}}:
Convolutional neural networks on graphs with fast localized spectral filtering.
Advances in neural information processing systems
\textbf{29}
(2016)
\end{botherref}
\endbibitem

\bibitem[\protect\citeauthoryear{Kipf and Welling}{2016}]{kipf2016semi}
\begin{botherref}
\oauthor{\bsnm{Kipf}, \binits{T.N.}},
\oauthor{\bsnm{Welling}, \binits{M.}}:
Semi-supervised classification with graph convolutional networks.
arXiv preprint arXiv:1609.02907
(2016)
\end{botherref}
\endbibitem

\bibitem[\protect\citeauthoryear{Scarselli et~al.}{2008}]{scarselli2008graph}
\begin{barticle}
\bauthor{\bsnm{Scarselli}, \binits{F.}},
\bauthor{\bsnm{Gori}, \binits{M.}},
\bauthor{\bsnm{Tsoi}, \binits{A.C.}},
\bauthor{\bsnm{Hagenbuchner}, \binits{M.}},
\bauthor{\bsnm{Monfardini}, \binits{G.}}:
\batitle{The graph neural network model}.
\bjtitle{IEEE transactions on neural networks}
\bvolume{20}(\bissue{1}),
\bfpage{61}--\blpage{80}
(\byear{2008})
\end{barticle}
\endbibitem

\bibitem[\protect\citeauthoryear{Hamilton et~al.}{2017}]{hamilton2017inductive}
\begin{botherref}
\oauthor{\bsnm{Hamilton}, \binits{W.}},
\oauthor{\bsnm{Ying}, \binits{Z.}},
\oauthor{\bsnm{Leskovec}, \binits{J.}}:
Inductive representation learning on large graphs.
Advances in neural information processing systems
\textbf{30}
(2017)
\end{botherref}
\endbibitem

\bibitem[\protect\citeauthoryear{Xu et~al.}{2018}]{xu2018powerful}
\begin{botherref}
\oauthor{\bsnm{Xu}, \binits{K.}},
\oauthor{\bsnm{Hu}, \binits{W.}},
\oauthor{\bsnm{Leskovec}, \binits{J.}},
\oauthor{\bsnm{Jegelka}, \binits{S.}}:
How powerful are graph neural networks?
arXiv preprint arXiv:1810.00826
(2018)
\end{botherref}
\endbibitem

\bibitem[\protect\citeauthoryear{Battaglia
  et~al.}{2018}]{battaglia2018relational}
\begin{botherref}
\oauthor{\bsnm{Battaglia}, \binits{P.W.}},
\oauthor{\bsnm{Hamrick}, \binits{J.B.}},
\oauthor{\bsnm{Bapst}, \binits{V.}},
\oauthor{\bsnm{Sanchez-Gonzalez}, \binits{A.}},
\oauthor{\bsnm{Zambaldi}, \binits{V.}},
\oauthor{\bsnm{Malinowski}, \binits{M.}},
\oauthor{\bsnm{Tacchetti}, \binits{A.}},
\oauthor{\bsnm{Raposo}, \binits{D.}},
\oauthor{\bsnm{Santoro}, \binits{A.}},
\oauthor{\bsnm{Faulkner}, \binits{R.}}, et al.:
Relational inductive biases, deep learning, and graph networks.
arXiv preprint arXiv:1806.01261
(2018)
\end{botherref}
\endbibitem

\end{thebibliography}

\end{document}